\documentclass[runningheads]{llncs}
\usepackage{graphicx}
\usepackage{hyperref}       % hyperlinks
\usepackage{url}            % simple URL typesetting
\usepackage{booktabs}       % professional-quality tables
\usepackage{amsfonts}       % blackboard math symbols
\usepackage{nicefrac}       % compact symbols for 1/2, etc.
\usepackage{microtype}      % microtypography
\usepackage{amsmath}
\usepackage{array}
\usepackage{booktabs}
\usepackage{dsfont}
\usepackage{graphicx}
\usepackage[table,xcdraw]{xcolor}
\usepackage[font=small,labelfont=bf]{caption}

\usepackage[colorinlistoftodos]{todonotes} % for comments

\DeclareMathOperator*{\argmax}{arg\,max}
\everypar{\looseness=-1}

\begin{document}
\title{PIMMS: Permutation Invariant Multi-Modal Segmentation}
\author{Thomas Varsavsky$^{1}$, Zach Eaton-Rosen$^{1,2}$, Carole H. Sudre$^{2,1,3}$,
Parashkev Nachev$^{4}$ \and M. Jorge Cardoso$^{2,1}$}
\institute{
CMIC, University College London, UK \\
%\email{\{thomas.varsavsky.15,z.eaton-rosen,carole.sudre.12,p.nachev\}@ucl.ac.uk}
\and
School of Biomedical Engineering and Imaging Sciences, Kings College London, UK \\
\and
Dementia Research Centre, University College London, UK \\
\and
Institute of Neurology, University College London, UK
%\email{\{m.jorge.cardoso\}@kcl.ac.uk}
}
\maketitle              % typeset the header of the contribution
\begin{abstract}
In a research context, image acquisition will often involve a pre-defined static protocol and the data will be of high quality. If we are to build applications that work in hospitals without significant operational changes in care delivery, algorithms should be designed to cope with the available data in the best possible way. In a clinical environment, imaging protocols are highly flexible, with MRI sequences commonly missing appropriate sequence labeling (e.g. T1, T2, FLAIR). To this end we introduce PIMMS, a Permutation Invariant Multi-Modal Segmentation technique that is able to perform inference over sets of MRI scans without using modality labels. We present results which show that our convolutional neural network can, in some settings, outperform a baseline model which utilizes modality labels, and achieve comparable performance otherwise. \looseness=-1
\end{abstract}
\section{Introduction}

Over the years, public medical imaging datasets have emerged which enable researchers to benchmark the performance of their algorithms \cite{WMHChallenge}. Data is mostly acquired from patients who have volunteered to be part of a clinical research study and are subject to a strict study protocol. If the study involves the acquisition of Magnetic Resonance Imaging (MRI) scans, the study protocol might dictate the scanner choice as well as the acquisition parameters to be used \cite{GhafoorianKHUSL16}. In the real unconstrained clinical setting however, MRIs are more likely to be acquired from different machines under different acquisition protocols and parameters. There is no guarantee that a particular sequence will be available, no guarantee on the number of available modalities, no guarantee that modalities will be unique (e.g. same sequence acquired with different orientations and contrasts), and no guarantee that any of the modalities will be labeled appropriately for algorithmic use. If hospitals are to benefit from advances in neuroimaging, algorithms that can cope with this lack of available modalities are necessary. We argue that an algorithm which is to be deployed in this setting should have two key properties: 1)permutation invariance, i.e permuting the order of the input images should not affect the output and 2)robustness to missing modalities. To this end we propose a segmentation model, with neural networks as building blocks, which can learn with limited data and segment scans without MR modality labels. In this work we focus on the task of segmenting white matter hyperintensities (WMH).
In studies involving WMH segmentation the most common modalities are T1, T2 and T2-FLAIR which provide complementary information about the imaged tissue. Although T1 and T2 modalities are created from different underlying physical signals (longitudinal and transverse relaxation time respectively) the scans produced will almost always be a combination of both (hence the name %name T1-
attribute -weighted). By varying the acquisition parameters, such as the echo and relaxation times,
these underlying physical signals are observed in different proportions
\cite{fischl2004sequence}. Modality labels are a discrete approximation of a continuous acquisition parameter landscape and we use this as inspiration for the model we present. \looseness=-1

In order to address missing modalities, research has focused mostly on generative models where missing MRI scans are synthesized or imputed \cite{iglesias2013synthesizing} \cite{chartsias2017multimodal}. In the work of \cite{havaei2016hemis} the authors handle missing modalities without using generative models of MR modalities. Instead of synthesizing the missing modalities, their model, Hetero-modal Image Segmentation (HeMIS), is trained to handle missing input modalities. More details about HeMIS can be found in section \ref{hemis}. Although HeMIS is successful at dealing with missing modalities, it assumes that the MR modalities in a test case will be labeled. %COMMENT: maybe use identified instead of labeled as you may want to reserve the word label for the segmentation.
The authors of \cite{karani2018lifelong} tackle the issue of generalizing to unseen protocols and scanners. In order to be robust to different scanners and protocols, they propose a tuning of the batch normalization parameters of a CNN.
%They propose a method to be robust to different scanners and protocols by tuning batch normalization parameters of a CNN. 
However, their method still requires approximately four scans with their associated segmentations from the unseen protocol to perform well. \looseness=-1

We introduce a model that learns to build intermediate representations of the %scans
images as a linear combination of the available inputs which are more continuous than their original labels. The proposed model does not assume the modality is known and has the ability to generalize to unseen scanners/protocols, taking in $N$ unordered input scans with no modality labels to produce accurate segmentation masks. 
We provide results on a variety of datasets featuring WMH with large variability in scanner type and acquisition parameters and show that our model is both permutation invariant and robust to missing modalities.
We demonstrate that it can perform comparatively well with an algorithm which utilizes the modality labels having never seen an image from that particular protocol. Furthermore, our model can outperform the baseline method (HeMIS) in the case where it has seen MR modality labels of the same protocol it is being tested on. \looseness=-1%COMMENT what is the baseline model? By labels do you mean segmentations or modalities?

\section{Methods}

\subsubsection{HeMIS}
\label{hemis}

 In HeMIS each available modality, $x_1, \ldots ,x_M$,  is embedded with a modality specific function $\phi_m(x_m) \in \mathbb{R}^{D \times K}$ denoted the ``back-end" to produce embeddings. An ``abstraction" layer then operates on these embeddings by computing the mean and variance across their $K$ dimensions and concatenating the two resulting vectors $
 \phi_\alpha = [\hat{\mathrm{E}}(\boldsymbol{\phi}(\textbf{x})), \hat{\mathrm{Var}}(\boldsymbol{\phi}(\textbf{x}))]
 $, where \textbf{x} $ \in \mathbb{R}^{D \times M}$ M is the number of modalities and D is the spatial dimensions of the input.
Let $\phi_\alpha$ be a fixed dimensional tensor which represents an input of variable size. This forms the input to the final portion of the network referred to as the ``frontend" which will output a semantic segmentation map. The network is trained using a Dice loss, first proposed in \cite{milletari2016v} as a loss function for training neural networks.

% In the case of binary segmentation this loss is defined as,% Get rid of this and fix references to it.
% \begin{equation}
% \mathcal{L}_{dice} = \frac{2 \sum^{N}_{i} \hat{y}_i y_i}{\sum^{N}_{i} \hat{y}^{2}_{i} + \sum^{N}_{i} y^{2}_{i}},
% \label{eq:dice}
% \end{equation}
% \noindent where $N$ is the voxels or pixels, $y$ is the ground truth and $\hat{y}$ is the prediction.
During training, random modalities are set to zero, encouraging robustness to missing modalities. HeMIS, shown in Figure \ref{network_architecture}, forms part of our architecture.

\subsubsection{Our approach}
\label{sec:our_approach}
We propose a method which at test time takes in an arbitrary number of $N$ scans (denoted $X$) which do not have corresponding MR modality labels and produces a permutation invariant representation that is also robust to missing modalities. % consider shortening.
In theory this common representation could be applied to a variety of tasks. In this paper we focus on white matter hyperintensity segmentation. \looseness=-1

The inputs are fed into an MR modality classifier $f_{mod}$ which outputs a distribution over modalities for a given scan as its prediction. These modality scores $\mathcal{S} \in \mathbb{R}^{M \times N}$ are combined with the inputs, $X$, to produce modified inputs denoted as $\hat{X} \in \mathbb{R}^{D \times M}$. In the attention literature a distinction is drawn between ``soft" and ``hard" attention \cite{xu2015show}. Soft attention generally involves a probabilistic weighted sum whilst a hard attention is a categorical argmax over the inputs. With this in mind, we explore two methods for performing $X \to \hat{X}$: $f_{soft}$ and $f_{hard}$. The function $f_{soft}$ is defined as,
\begin{equation}
f_{soft}(X, \mathcal{S}) = \sum_{n=1}^{N} \mathcal{S}_{mn} x_{n}  = \hat{x}_{m}
\end{equation}
Each component $\hat{x}_{m}$ of the modified input $\hat{X}$ is formed by taking a weighted sum of each input $x_{n}$ according to the probabilities provided by $\mathcal{S}$. $f_{hard}$ is defined as,

\begin{equation}
f_{hard}(X, \mathcal{S}) = \sum_{n=1}^{N} \mathds{1}(\argmax_{m^{*}} \mathcal{S}_{m^{*}n} = m) x_n = \hat{x}_{m}
\end{equation}

\begin{figure}[h]
\centering
\includegraphics[width=\textwidth]{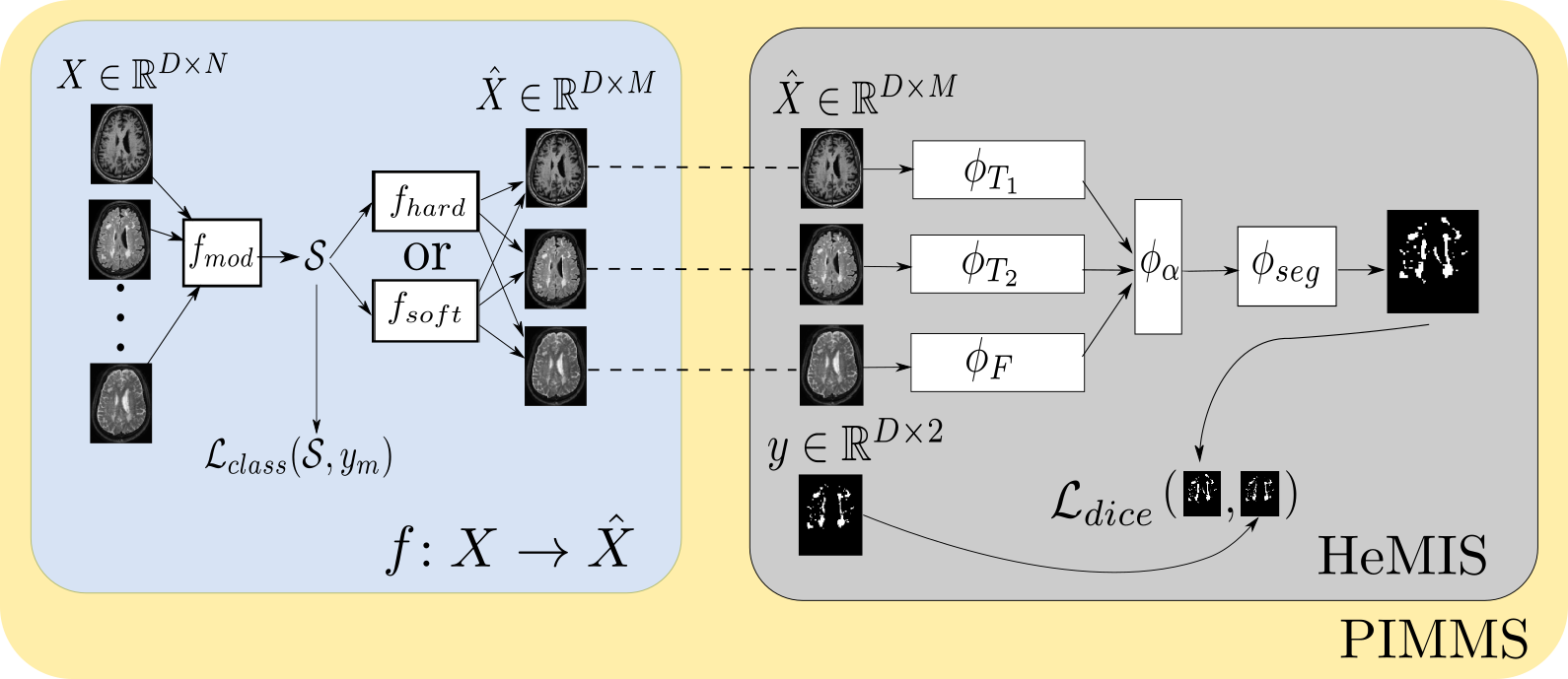}
\caption{Diagram showing the network architecture. During training the inputs are $X \in \mathbb{R}^{D \times N}$ and the corresponding ground truth binary segmentation $y \in \mathbb{R}^{D \times 2}$. A function $f_{mod}$ takes each scan as input and outputs a modality score $\mathcal{S}$ which produces the representation $\hat{X} \in \mathbb{R}^{D \times M}$. The weights of $\phi_{T_1}$, $\phi_{T_2}$, $\phi_{F}$ and $\phi_{seg}$ are learned by differentiating with respect to $\mathcal{L}_{seg}$ and the weights of $f$ are learned by differentiating with respect to $\mathcal{L}_{class}$. $y_m$ is a one-hot encoded modality label.}
\label{network_architecture}
\vskip -15pt
\end{figure}

The modified input $\hat{X}$ now consists of a finite number of modalities. The mapping $f: X \to \hat{X}$ is illustrated in the blue block in Figure \ref{network_architecture}.

Each MR modality is designed to capture fundamentally different physical properties which justifies having individual feature extractors, $\phi_m$, for each $\hat{x}_m$ modality representation. The output of these modality-specific feature extractors is collected into one tensor by taking the mean and the variance across modalities and concatenating the result to give $\phi_\alpha \in \mathbb{R}^{D \times K}$ where $K$ is given by the choice of filter depth in $\phi_m$. This feeds into a final network, $\phi_{seg}$ which produces a segmentation prediction. This use of modality specific models, pooling and a separate segmentation network is the same as HeMIS and is illustrated in the grey block in Figure \ref{network_architecture}. \looseness=-1

% MENTION THAT HARD CONVERGES TO HEMIS AS ACCURACY TENDS TO 1
 A convolutional neural network was used for $f_{mod}$. A network with 36 layers using skip connections and ReLU non-linearities inspired by the residual network (ResNet) proposed in \cite{he2016identity} is used. The network was trained with the categorical cross-entropy loss which we refer to as $\mathcal{L}_{class}$.
where $y_{mi}$ is a one-hot encoded modality label and $\mathcal{S}_{mi}$ is the modality score. Each of the branches $\phi_{m}$ as well as $\phi_{seg}$ were two convolutional layers with ReLU non-linearities (more details in section \ref{implementation}). The parameters of $\phi_{m}$ and $\phi_{seg}$ were found by minimizing $\mathcal{L}_{seg}$ which is the binary Dice Loss. 

For two of our variants these losses were trained separately (or ``offline"). However, we also trained an ``online" variant where the parameters of the modality classifier are learned using a multi-objective loss function. This loss is defined as, $\mathcal{L}_{tot} = \mathcal{L}_{seg} + \lambda \mathcal{L}_{class} $, where $\lambda$ is some choice of weighting or parametrized weighting function. Although the loss consists of multiple objectives this should not be considered ``multi-task learning". There is no conditional independence between the tasks and no representation sharing --- instead this can be seen as a differentiable attention mechanism. The four variants trained are summarized below, \looseness=-1
\label{model_description}
\begin{description}
\item [HeMIS] - $X \to \hat{X}$ using labels, $f_{mod}$ trained separately  from $\phi_{seg}$, $\phi_{T_1}$, $\phi_{T_2}$ \& $\phi_{F}$ \looseness=-1
\item [Soft] - $f_{soft}$ used to create $\hat{X}$, $f_{mod}$ trained separately  from $\phi_{seg}$, $\phi_{T_1}$, $\phi_{T_2}$ \& $\phi_{F}$,\looseness=-1
\item [Hard] - $f_{hard}$ used to create $\hat{X}$,  $f_{mod}$ trained separately  from $\phi_{seg}$, $\phi_{T_1}$, $\phi_{T_2}$ \& $\phi_{F}$, \looseness=-1
\item [Online] - $f_{soft}$ used to create $\hat{X}$, $f_{mod}$ trained jointly with $\phi_{seg}$, $\phi_{T_1}$, $\phi_{T_2}$ \& $\phi_{F}$.\looseness=-1,
\end{description}
\vspace{-7mm}
\subsubsection{Implementation Details} \label{implementation}

It is important to note that the network architecture takes in 2D patches from the image as was done in \cite{havaei2016hemis}. Specifically we take patches of size $100 \times 100$ from 3D scans which have all been resampled to $1mm \times 1mm \times 1mm$. This theoretical framework permits any spatial dimension $D$ and future work will train and run inference in full 3D.

All results were obtained using the NiftyNet framework \cite{gibson2017niftynet}, which is a wrapper around TensorFlow designed for medical imaging. $f_{mod}$ uses a standard ResNet design with nine blocks per resolution, each with three convolutions and relu activations. The network is trained using the Adam optimizer with a learning rate of $3 \times 10^{-4}$. A batch size of $64$ was used on this network and weight decay regularization of $1 \times 10^{-4}$. 

For each $\phi_{m}$ and $\phi_{seg}$ the implementation details from \cite{havaei2016hemis} were recreated. Two convolutional layers with 48 filters, $5 \times 5$ kernel sizes, zero-padding and ReLU activation were used followed by a max pooling layer with kernel size (2, 2) and a stride of 1 this preserves the spatial resolution of the image. For $\phi_{seg}$ two convolutional layers were used, one with 16 filters, $5 \times 5$ kernel sizes, zero padding and ReLU activation the last convolutional layer had 2 filters, a kernel size of $21 \times 21$, zero padding and a softmax activation which provided the per class predictions. We also utilized the pseudo-curriculum learning approach from HeMIS. Random modalities are set to zero but the chance of setting only one or no modalities to zero is higher. The online model was harder to train than the offline ones. The joint training lead to odd dynamics between the classification loss and the segmentation loss. To help stabilize the training an exponential decay weighting was used on the classification loss in order to encourage training it towards the start and remove its importance later on so that the model could experiment with representations which do not match the provided labels and not be punished by $\mathcal{L}_{class}$. Our best performing ``online" model used $\lambda(i) = e^{-\gamma i}$ where $i$ is the current iteration and $\gamma$ is a decay constant hyperparameter set to $1 \times 10^{-4}$. \looseness=-1

This same ResNet architecture was used as $f_{mod}$ in the online case in order to make a fair comparison in terms of number of parameters. However, in the online setting, the batch size had to be reduced as a practical consideration as the combination of both modality and backend models proved too large to fit in GPU memory. All experiments were run on a single NVIDIA Titan Xp.

\section{Experiments and Results}
\label{sec:experiments}
% We have thus far proposed a method that would provide permutation invariant representations of the inputs in terms of a finite number of $\hat{x}_m$ modality representations, through the use of some function $f: X \to \hat{X}$. In this section we present results for \\
%\subsubsection{Data} 
Data used in this work comes from a variety of sources, chosen to try and capture the acquisition variability observed in a practical setting due to multiple MRI scanners/protocols.  A subset of 973 subjects each with T1 and FLAIR scans were obtained from the Alzheimer's Disease Neuroimaging Initiative (ADNI) database \cite{jack2008alzheimer}. The data in this study was collected from multiple scanners, but used the same protocol for setting the acquisition parameters. We therefore deem this dataset one of relatively low variance between subjects. We also utilise data collected from the longitudinal SABRE study \cite{tillin2010southall}. The data contains one cohort of 586 subjects with T1, T2 and FLAIR obtained using the same scanner (low variance) and another of 1263 with T1, T2 and FLAIR obtained from multiple scanners with multiple settings (high variance). Additionally we use a dataset of 626 patients with T1 and FLAIR obtained from multiple scanners using multiple field strengths. As no manual annotations were available for this large collection of MRI scans, the outputs of BaMoS \cite{sudre2015bayesian}, a fully unsupervised WM lesion segmentation algorithm, were quality controlled by an experienced human rater and subsequently used as silver-standard training labels. Additionally, we evaluate our trained models on a manually annotated dataset from the MICCAI 2017 White Matter Hyperintensity Challenge \cite{WMHChallenge}.

%\subsubsection{Results} 
The split between training, validation and test sets was chosen in order to measure the ability of our method at generalizing to unseen scanners and protocols. Three separate holdouts were created, defined as follows,
\begin{description}
\item [Mixed Holdout] - Random 10\% subset of the full data minus ADNI/MICCAI17,
\item [Silver Protocol Holdout] - ADNI: 973 subjects with silver standard labels.
\item [Gold Protocol Holdout] - MICCAI2017: 60 subjects with human rater labels.
\end{description}
Overall there was a 80/10/10 split between training, validation and test using the 2474 subjects that are not in the gold or silver protocol holdouts. All four models described in section \ref{model_description} were trained with this subset.

\begin{figure*}[h!]
\centering
\begin{tabular}{cccccc}
\multicolumn{2}{c}{95\%} & \multicolumn{2}{c}{50\%} & \multicolumn{2}{c}{20\%} \\
GT & Pred & GT & Pred & GT & Pred \\
\includegraphics[width=0.16\linewidth, trim={55pt 35pt 35pt 40pt}, clip=true]{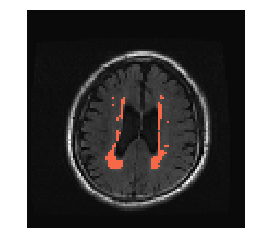} &
\includegraphics[width=0.16\linewidth, trim={55pt 35pt 35pt 40pt}, clip=true]{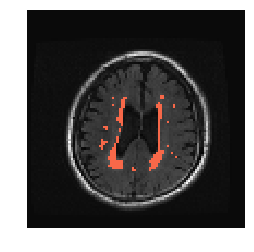} &
\includegraphics[width=0.16\linewidth, trim={45pt 40pt 30pt 18pt}, clip=true]{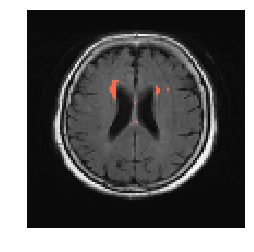} &
\includegraphics[width=0.16\linewidth, trim={45pt 40pt 30pt 18pt}, clip=true]{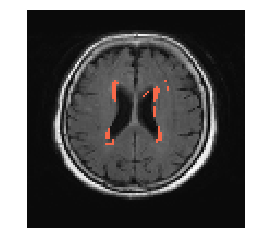} &
\includegraphics[width=0.16\linewidth, trim={55pt 36pt 28pt 30pt}, clip=true]{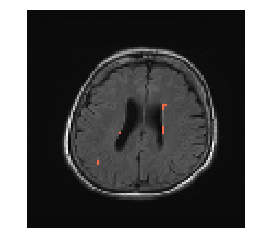} &
\includegraphics[width=0.16\linewidth, trim={55pt 36pt 28pt 30pt}, clip=true]{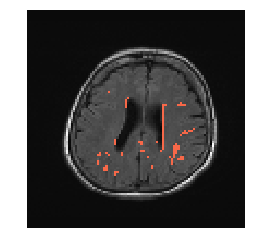}
\end{tabular}
% COMMENT: It was 5th percentile before and you say 20th percentile in the caption...
\caption{Qualitative results showing white matter lesion segmentations on the mixed holdout set. Images show the ground truth on the left and the network predictions on the right. Red shows the predicted segmentation. The results were chosen to highlight the 95th, 50th, and 20th percentile in terms of Dice score for a model which is trained on all available scans but does not use modality labels.}
\label{qualitative}
\vskip -15pt
\end{figure*}

\newcolumntype{M}[1]{>{\centering\arraybackslash}m{#1}}
\newcolumntype{N}{@{}m{0pt}@{}}
\begin{table}[h]
\centering
\makebox[0pt][c]{\parbox{1.106\textwidth}{
\setlength\tabcolsep{4pt}
\begin{minipage}{0.555\textwidth}
\resizebox{\textwidth}{!}{
\centering
\begin{tabular}{ccc|cccccccc|}
\cline{4-11}
\multicolumn{3}{l|}{} & \multicolumn{8}{c|}{Mixed Holdout} \\ \hline
\multicolumn{3}{|c|}{Modalities} & \multicolumn{4}{c|}{Dice Score} & \multicolumn{4}{c|}{Avg. Symmetric Distance} \\ \hline
\rowcolor[HTML]{808080}
\multicolumn{1}{|c|}{$T_{1}$}  & \multicolumn{1}{c|}{$T_{2}$} & $F$ & \multicolumn{1}{c|}{HeMIS}  & \multicolumn{1}{c|}{Soft} & \multicolumn{1}{c|}{Hard} & \multicolumn{1}{c|}{Online} & \multicolumn{1}{c|}{HeMIS}  & \multicolumn{1}{c|}{Soft} & \multicolumn{1}{c|}{Hard} & \multicolumn{1}{c|}{Online} \\ \hline

\rowcolor[HTML]{C0C0C0} 
\multicolumn{1}{|c}{$\bullet$} & $\bullet$ & $\bullet$ & 0.47 & \textbf{0.51} & 0.48 & \textbf{0.54} & 0.71 & 0.65 & 0.71 & 1.9\\
\multicolumn{1}{|c}{\cellcolor[HTML]{C0C0C0}$\bullet$} & \cellcolor[HTML]{C0C0C0}$\bullet$ & \cellcolor[HTML]{C0C0C0}$\circ$ & 0.3 & \textbf{0.39} & 0.3 & 0.24 & 2.32 & \textbf{1.92} & 2.36 & 4.21\\
\rowcolor[HTML]{C0C0C0}
\multicolumn{1}{|c}{$\circ$} & $\bullet$ & $\bullet$ & 0.26 & \textbf{0.32} & \textbf{0.26} & \textbf{0.4} & 0.77 & 0.82 & 0.76 & 3.32\\
\multicolumn{1}{|c}{\cellcolor[HTML]{C0C0C0}$\bullet$} & \cellcolor[HTML]{C0C0C0}$\circ$ & \cellcolor[HTML]{C0C0C0}$\bullet$ & 0.44 & \textbf{0.45} & \textbf{0.45} & \textbf{0.52} & 0.61 & 0.63 & 0.62 & 2.06\\
\rowcolor[HTML]{C0C0C0} 
\multicolumn{1}{|c}{$\bullet$} & $\circ$ & $\circ$ & 0.1 & \textbf{0.1} & 0.1 & \textbf{0.19} & 3.42 & 3.76 & 3.51 & 4.48\\
\multicolumn{1}{|c}{\cellcolor[HTML]{C0C0C0}$\circ$} & \cellcolor[HTML]{C0C0C0}$\bullet$ & \cellcolor[HTML]{C0C0C0}$\circ$ & 0.08 & \textbf{0.08} & 0.07 & \textbf{0.09} & 4.07 & 4.13 & 4.53 & 7.48\\
\rowcolor[HTML]{C0C0C0}
\multicolumn{1}{|c}{$\circ$} & $\circ$ & $\bullet$ & 0.16 & \textbf{0.18} & \textbf{0.16} & \textbf{0.41} & 0.56 & 0.61 & \textbf{0.54} & 3.31\\
\end{tabular}
}
\end{minipage}%
\begin{minipage}{0.54\textwidth}
\resizebox{\textwidth}{!}{
\centering
\begin{tabular}{ccc|cccccccc|}
\cline{1-11}
\multicolumn{3}{|c|}{Modalities} & \multicolumn{4}{c|}{Dice Score} & \multicolumn{4}{c|}{Avg. Symmetric Distance} \\ \hline
\rowcolor[HTML]{808080}
\multicolumn{1}{|c|}{$T_{1}$}  & \multicolumn{1}{c|}{$T_{2}$} & $F$ & \multicolumn{1}{c|}{HeMIS}  & \multicolumn{1}{c|}{Soft} & \multicolumn{1}{c|}{Hard} & \multicolumn{1}{c|}{Online} & \multicolumn{1}{c|}{HeMIS}  & \multicolumn{1}{c|}{Soft} & \multicolumn{1}{c|}{Hard} & \multicolumn{1}{c|}{Online} \\ \hline
\cline{1-11}
\multicolumn{11}{|c|}{Silver Protocol Holdout} \\ \hline

\rowcolor[HTML]{C0C0C0} 
\multicolumn{1}{|c}{$\bullet$} & $\circ$ & $\bullet$ & 0.48 & 0.46 & 0.46 & 0.44 & 0.68 & 0.72 & 1.12 & 3.52\\
\multicolumn{1}{|c}{\cellcolor[HTML]{C0C0C0}$\bullet$} & \cellcolor[HTML]{C0C0C0}$\circ$ & \cellcolor[HTML]{C0C0C0}$\circ$ & 0.11 & 0.11 & 0.08 & \textbf{0.21} & 0.79 & 0.79 & 1.63 & 5.17\\
\rowcolor[HTML]{C0C0C0}
\multicolumn{1}{|c}{$\circ$} & $\circ$ & $\bullet$ & 0.25 & 0.16 & 0.24 & \textbf{0.5} & 0.69 & \textbf{0.68} & 0.8 & 2.77\\
\cline{1-11}
 \multicolumn{11}{|c|}{Gold Protocol Holdout} \\ \hline

\rowcolor[HTML]{C0C0C0} 
\multicolumn{1}{|c}{$\bullet$} & $\circ$ & $\bullet$ & 0.59 & 0.64 & 0.62 & 0.61 & 0.76 & \textbf{0.57} & 0.72 & 1.18\\
\multicolumn{1}{|c}{\cellcolor[HTML]{C0C0C0}$\bullet$} & \cellcolor[HTML]{C0C0C0}$\circ$ & \cellcolor[HTML]{C0C0C0}$\circ$ & 0.41 & 0.35 & 0.42 & \textbf{0.47} & 0.8 & \textbf{0.44} & \textbf{0.77} & 2.18\\
\rowcolor[HTML]{C0C0C0}
\multicolumn{1}{|c}{$\circ$} & $\circ$ & $\bullet$ & $0.38$ & $0.38$ & $0.26$ & $\textbf{0.45}$ & $1.01$ & $1.75$ & $20.75$ & $3.63$\\
\end{tabular}}
\end{minipage}%
}}

\vspace{1em}
\caption{Dice scores of the different models on different combinations of available modalities. Modalities present are denoted by $\bullet$ and those that are missing are denoted by $\circ$. Bold numbers are results which outperform the baseline model, HeMIS, with statistical significance $p<0.01$ as provided by a Wilcoxon test.º Presentation of table inspired by the one in \cite{havaei2016hemis} \looseness=-1}
\label{table:main}
\vskip -15pt
\end{table}
For the mixed holdout it was found that the classification accuracy was $99\%$ between all three modalities. For unseen protocols the accuracy was lower, $88\%$ for ADNI and $87\%$ for MICCAI17 which showed that the inter-scanner variance was harder to model than the inter-subject variance. 
For each of the holdout sets, results are presented on all possible subsets of the available modalities. The quantitative and qualitative results are shown in Table \ref{table:main} and Figure \ref{qualitative}, respectively. The brains shown are selected from the 95\%, 50\% and 20\% percentile of Dice score on the dataset holdout for a model shown all available modalities. We note that the samples of very high Dice score are often the ones with large lesions which the algorithm has managed to capture well and there is poor performance when the contrast settings are significantly different. \looseness=-1

We utilise the Wilcoxon signed-rank test to test whether the Dice scores from each of our models outperforms the baseline (HeMIS). Bold values in Tables \ref{table:main} denotes that the model is better than HeMIS with a statistical significance of $p<0.01$. We compare ground truths and predictions using the Dice score as well as the average symmetric distance in order to provide a geometric evaluation. \looseness=-1

\section{Discussion}
The ``hard" setting converges to HeMIS as the accuracy of the modality classifier tends to 1. This is observed in practice. Note that the results of HeMIS are similar to ``hard" in the mixed holdout set where the modality classifier has had access to the test set distribution and consistently worse in the Silver Protocol holdout. It does comparatively better on the Gold Protocol as the modality classifier has better performance on these scans than on Silver. The ``soft" version matches or improves on the performance of HeMIS and ``hard" on the mixed holdout, but does not outperform HeMIS on other holdouts. The fact that ``soft" outperfoms ``hard" is evidence towards our hypothesis that mixing the input images can lead to better representations which improve performance on a visual task. This can be interpreted as a coarse attention mechanism as the transformation from $X$ to $\hat{X}$ is linear with few degrees of freedom. \looseness=-1

The ``online" model outperforms the baseline in the mixed holdout set with statistical significance in 6/7 cases when using the Dice score. Although the median average symmetric distance (ASD) is higher, the average is lower in 4/7 cases with a much lower 95 percentile. There is some improvement over the baseline model even in the protocol holdout but the gains seen in Dice score are not reflected in the ASD. Qualitatively this is explained by the ``online" method overpredicting the positive class leading to a higher Dice score and yet missing lesions altogether leading to a larger ASD. This gives us insights as to how we can improve the model. \looseness=-1

Future work will extend the ``online" model to an unsupervised setting in terms of scan labels. This is appealing not only due to the lack of modality labels currently available in certain hospital databases but also in order to go \textit{beyond} the information contained in the modality label and towards a representation which is more true to the underlying physical structure. \looseness=-1
\section{Conclusion}
We have presented PIMMS, a segmentation algorithm for MRI scans which simultaneously addresses the problem of missing modalities and missing modality labels in a clinical setting. We present three variants which all include a convolutional neural network and are trained to perform modality classification in a supervised setting. 
% We argue that by combining the input modalities with learned mixing proportions
We argue that by mixing the input modalities in ratios other than those provided by the labels we can achieve better performance. This could be due to more accurately capturing the underlying distribution of physical quantities, but future work is needed to make this claim. Evidence is presented with statistical significance which suggests that a model which mixes inputs can perform better than one which does not with all other factors kept identical.

The results show that the modality classifier almost replicates modality labels when trained and tested on the same protocol while the categorical accuracy reaches 88\% when protocols differ at training and testing times.
%The results presented show that a modality classifier can be learned that almost exactly replicates the modality labels when the model has trained on the same protocol as test and achieves a categorical accuracy of 88\% when it has not. 
Our model serves as a proof of concept for a system that could utilize all the MR scans associated with a patient in a hospital and provide accurate segmentation predictions. \looseness=-1 \\

\textit{Acknowledgements} We gratefully acknowledge the support of NVIDIA Corporation with the donation of the Titan Xp GPU used for this research. Zach Eaton-Rosen is supported by the EPSRC Doctoral Prize. Carole H. Sudre is supported by the Biomedical Junior Fellowship from Alzheimer’s Society. Parashkev Nachev is funded by the Wellcome Trust and the UCLH NIHR Biomedical Research Centre. M. Jorge Cardoso is funded by Wellcome Trust.

% \section{Acknowledgments}
% We gratefully acknowledge the support of NVIDIA Corporation with the donation of the Titan Xp GPU used for this research. Zach Eaton-Rosen is supported by the EPSRC Doctoral Prize. Carole H. Sudre is supported by the Biomedical Junior Fellowship from Alzheimer’s Society. Parashkev Nachev is funded by the Wellcome Trust, the Department of Health, and the UCLH NIHR Biomedical Research Centre. M. Jorge Cardoso is funded by Wellcome grant: HICF-R9-501.
\bibliographystyle{splncs04}
\bibliography{main.bbl}
%\bibliography{biblio}
\end{document}